\documentclass[conference]{IEEEtran}
\IEEEoverridecommandlockouts
\usepackage{cite}
\usepackage{amsmath,amssymb,amsfonts}
\usepackage{algorithmic}
\usepackage{graphicx}
\usepackage{textcomp}
\usepackage{xcolor}
\usepackage{booktabs}
\usepackage{multirow}
\usepackage[table]{xcolor}
\usepackage{array}
\usepackage{caption}
\usepackage{makecell}

\def\BibTeX{{\rm B\kern-.05em{\sc i\kern-.025em b}\kern-.08em
    T\kern-.1667em\lower.7ex\hbox{E}\kern-.125emX}}
\begin{document}

\title{Learning to Retrieve Navigable Candidates for Efficient Vision-and-Language Navigation}

\author{
\IEEEauthorblockN{
Shutian Gu\textsuperscript{1},
Chengkai Huang\textsuperscript{1,2},
Ruoyu Wang\textsuperscript{1},
Lina Yao\textsuperscript{1,3}
}
\IEEEauthorblockA{
\textsuperscript{1}University of New South Wales, Sydney, Australia \\
\textsuperscript{2}Macquarie University, Sydney, Australia \\
\textsuperscript{3}Data61, CSIRO, Sydney, Australia \\
Email: chengkai.huang1@unsw.edu.au
}
}

% \author{\IEEEauthorblockN{Anonymous Authors}}

\maketitle

\begin{abstract}
Vision-and-Language Navigation (VLN) requires an agent to follow natural-language instructions and navigate through previously unseen environments. Recent approaches increasingly employ large language models (LLMs) as high-level navigators due to their flexibility and reasoning capability. However, prompt-based LLM navigation often suffers from inefficient decision-making, as the model must repeatedly interpret instructions from scratch and reason over noisy and verbose navigable candidates at each step. In this paper, we propose a retrieval-augmented framework to improve the efficiency and stability of LLM-based VLN without modifying or fine-tuning the underlying language model. Our approach introduces retrieval at two complementary levels. At the episode level, an instruction-level embedding retriever selects semantically similar successful navigation trajectories as in-context exemplars, providing task-specific priors for instruction grounding. At the step level, an imitation-learned candidate retriever prunes irrelevant navigable directions before LLM inference, reducing action ambiguity and prompt complexity. Both retrieval modules are lightweight, modular, and trained independently of the LLM. We evaluate our method on the Room-to-Room (R2R) benchmark. Experimental results demonstrate consistent improvements in Success Rate, Oracle Success Rate, and SPL on both seen and unseen environments. Ablation studies further show that instruction-level exemplar retrieval and candidate pruning contribute complementary benefits to global guidance and step-wise decision efficiency. These results indicate that retrieval-augmented decision support is an effective and scalable strategy for enhancing LLM-based vision-and-language navigation.
\end{abstract}


\section{Introduction}

Vision-and-Language Navigation (VLN) requires an agent to follow natural language instructions and navigate through discrete viewpoints in a previously unseen environment \cite{AndersonWTB0S0G18}. 
The task demands accurate grounding of linguistic cues in visual observations as well as sequential decision-making over long horizons.
Recent advances increasingly adopt large language models (LLMs) as high-level navigators \cite{zhou2023navgpt,huang2025pluralistic}, formulating VLN as a sequence of language-based reasoning and action selection steps \cite{ahn2022icanisay, huang2022innermonologue, zhao2024expel}.
In LLM-based VLN frameworks, the navigator is typically modeled as a language-driven decision module that predicts navigation actions from textual descriptions of the current observation, navigation history, and navigable candidates.
This formulation allows LLMs to naturally incorporate long-range instruction context and produce interpretable reasoning traces in natural language \cite{wei2022chain,wang2026scenealign}, making LLM-based navigation a promising direction for embodied decision making. 
By leveraging the strong reasoning ability and world knowledge encoded in LLMs \cite{brown2020language, touvron2023llama, yang2025qwen3,ye2026memweaver,jiao2026prunerag}, these methods enable more flexible decision making and improved generalization, particularly in unseen environments \cite{wang2024learning, zhao2024expel}.

Despite these strengths, prompt-based navigation with LLMs exhibits notable limitations.
In particular, the core challenge no longer lies in visual perception, but in how decisions are made under complex and information-heavy prompts.

\textbf{\textit{Gap 1: Lack of task-specific priors for instruction understanding.}}
At the beginning of each navigation episode, the LLM is typically required to interpret the instruction and infer appropriate navigation strategies from scratch.
Although similar instructions and successful navigation patterns may exist in prior experiences, current approaches do not explicitly leverage such knowledge, placing unnecessary burden on the LLM's reasoning process \cite{wei2022chain, wang2024learning,ye2026memweaver}.

\textbf{\textit{Gap 2: Inefficient and noisy candidate-level decision making.}}
At each navigation step, the agent is exposed to a relatively large set of navigable candidates, each accompanied by verbose textual descriptions.
The LLM must repeatedly reason over all candidates to select the next action \cite{long2023discuss, zhou2023navgpt}, even though many of them are clearly suboptimal or irrelevant given the current state.
Requiring the LLM to reason over all candidates increases inference cost and susceptibility to decision errors, especially in unseen environments.

A key observation underlying our work is that successful navigation trajectories exhibit strong and reusable decision structure. At the episode level, prior successful trajectories encode effective instruction grounding and navigation strategies that can serve as useful exemplars for new tasks.
At the step level, navigation decisions are highly structured: in most states, only a small subset of candidates aligns with the optimal path, while the majority are clearly suboptimal.
Nevertheless, existing LLM-based VLN methods largely ignore these regularities, leaving both instruction understanding and candidate selection to the LLM’s implicit reasoning \cite{zhou2023navgpt, long2023discuss}.
As a result, task-relevant structure must be repeatedly rediscovered from raw prompts, leading to inefficient and unstable decisions.

Motivated by this observation, we propose a dual-level retrieval framework that augments LLM-based navigation with explicit decision support.
Before navigation begins, an instruction-level exemplar retriever selects a small set of semantically similar successful trajectories as in-context demonstrations.
During navigation, a candidate retriever scores and prunes navigable actions at each step, allowing the LLM to reason over a reduced candidate set.
Together, these retrieval modules reduce the reasoning burden on the LLM while preserving its flexibility and interpretability.
Both modules are lightweight, modular, and operate independently of the LLM.
In particular, the candidate retriever is trained via imitation learning with shortest-path supervision, without modifying or fine-tuning the navigator.

In this paper, our contribution can be summarized as follows:
\begin{itemize}
    \item We introduce an instruction-level exemplar retrieval mechanism for LLM-based vision-and-language navigation, enabling the reuse of successful navigation experiences as in-context guidance.
    \item We propose an imitation-learned candidate retriever that explicitly models action relevance and prunes suboptimal navigable candidates prior to LLM decision making.
    \item We demonstrate on the R2R benchmark that the proposed dual-level retrieval framework consistently improves navigation performance, decision stability, and efficiency, particularly in unseen environments.
    % decision stability under distribution shift, particularly in unseen environments.
\end{itemize}

\section{Related Work}

\subsection{Vision-Language Navigation}

Recent progress in multimodal learning and retrieval 
\cite{liu2021semi, wang2025self, li2021autoencoder, li2021self} 
has advanced vision-language representation modeling. 
Vision-Language Navigation (VLN) has received significant attention in recent years, and a wide range of approaches have been proposed to improve navigation performance and generalization.
Early VLN methods primarily rely on supervised learning with multimodal encoders \cite{qi2020objectaction, hong2021vlnbert, chen2021historyaware} and sequential decision models \cite{deng2020graphical, lin2022adapt}, often integrating local perception with short-term memory or global maps \cite{an2023bevbert, wang2023scalingdata}. While effective for well-annotated datasets, these approaches struggle with long-horizon reasoning and generalization to novel environments due to limited task priors and reliance on densely labeled trajectories.
Recently, LLM-based VLN has emerged as a promising direction, leveraging large language models to perform embodied decision-making through textual reasoning \cite{zhao2023dynamic, huang2022innermonologue}. These agents transform multimodal inputs into language representations and rely on LLM reasoning to plan action sequences, decompose sub-goals, and apply commonsense knowledge in zero-shot or few-shot settings \cite{wei2022chain}. 
% NavGPT is a representative example of this paradigm, constructing a purely LLM-based agent that uses textual observations and navigation history to make action decisions, demonstrating strong generalization and interpretable reasoning without task-specific training. 
Despite their flexibility, such methods largely depend on carefully designed prompts to encode visual observations and navigation history \cite{long2023discuss, zhou2023navgpt}, which limits the ability of the agent to learn task priors end-to-end and constrains robust long-horizon performance.

Following these mainstream studies, our work builds on the advantages of NavGPT \cite{zhou2023navgpt}, enhancing LLM-based VLN by introducing embedding and candidate retrievers to incorporate learned task knowledge, reduce reliance on prompt engineering, and improve reasoning fidelity and navigation performance in complex environments.

\subsection{In-Context Learning}

In-context learning (ICL) describes the capability of large language models to perform tasks by conditioning on a small set of input–output examples provided in the prompt, without parameter updates \cite{brown2020language, wei2021finetuned,wang2025federated,huang2025embedding}.
Prior studies in natural language processing show that demonstrations implicitly shape model behavior and reasoning trajectories \cite{liu2021makes, rubin2021learning,huang2025listwise}.
ICL performance is known to be sensitive to demonstration selection, ordering, and formatting, which has motivated research on prompt calibration and example selection \cite{liu2021makes, hao2022structured, liu2023context}.
Retrieval-augmented ICL further addresses this issue by dynamically selecting demonstrations based on input similarity or learned relevance, and has shown strong empirical results on static prediction tasks \cite{wei2022chain}.

Beyond single-step NLP settings, applying in-context learning to embodied and sequential decision-making problems introduces additional challenges \cite{yao2023react, kagaya2024rap}.
In such tasks, decisions are state-dependent and unfold over long horizons, making it insufficient to treat demonstrations as isolated input–output pairs.
Recent works have begun to explore ICL in agentic contexts, including embodied agents and navigation tasks, typically by incorporating a small number of handcrafted or heuristically selected exemplars into the prompt \cite{kagaya2024rap}.
However, these approaches often rely on fixed demonstrations or manual prompt design, and do not explicitly model how prior experiences should be selected or adapted to the current task and environment \cite{wang2024survey}.
In the context of vision-and-language navigation, where instruction grounding and action selection evolve throughout an episode, how to effectively leverage prior navigation experience as in-context guidance remains largely underexplored.

Our work builds on this line of research by introducing task-aware in-context learning for VLN.
By retrieving exemplar trajectories conditioned on the current instruction and integrating them into the navigation prompt, we enable LLM-based agents to reuse prior navigation experience in a structured manner while preserving the flexibility of language-driven reasoning.

\section{Problem Formulation}\label{sec:problem}

% Vision-and-Language Navigation (VLN) requires an agent to follow a language instruction $I$ to navigate from a start viewpoint to a target viewpoint.
% At timestep $t$, the agent receives a panoramic observation $O_t$ containing $K$ single-view observations $O_{t,k}$, i.e.,
% \[
% O_t = \{ O_{t,k} \}_{k=1}^{K}.
% \]
% Among the $K$ views, there are $N$ navigable views.
% The navigable views together with a stop token $\texttt{[stop]}$ form the action space, from which the agent selects one as the action prediction $a_t$.
% Actions before step $t$ are regarded as the navigation history, denoted as
% \[
% H_t = \{ a_0, \ldots, a_{t-1} \}.
% \]
% A navigation trajectory is considered successful when the agent stops within $3$ meters of the target viewpoint.

We study VLN in the Room-to-Room (R2R) setting \cite{AndersonWTB0S0G18}, where an agent must follow a natural-language instruction to navigate on a discrete viewpoint graph from a start viewpoint to a goal region.

\textbf{Instruction.} Each episode provides an instruction $I$ that includes the task instruction text and a fixed set of system navigation rules.

\textbf{Observation.} At step $t$, the agent is at the current viewpoint $v_t$ with a heading/elevation orientation. The environment provides a panoramic observation that we represent in text form as:
\begin{equation}
    o_t = \left( s_t^{scene}, \{d_{t,k}\}_{k=1}^{K} \right), \quad K=8,
\end{equation}

where $s_t$ is a short scene summary of the current viewpoint, and $d_{t,k}$ is the textual description for the $k$-th directional sector (Front, Front-Right, \dots, Front-Left). Each $d_{t,k}$ aggregates (i) a direction-specific scene description, (ii) nearby object mentions (within 3 meters), and (iii) navigable viewpoint IDs that fall into that direction sector with approximate relative heading and distance (as constructed by our directional binning function).

\textbf{Candidates.} We define the directional candidate set as:
\begin{equation}
\mathcal{D}_t = \{ d_{t,k} \}_{k=1}^{K},
\end{equation}

and denote by $\mathcal{V}_{t,k} \subseteq \mathcal{V}_t$ the subset of navigable viewpoints whose headings fall into sector $k$ (these are the viewpoint IDs listed inside $d_{t,k}$). Thus, each $d_{t,k}$ is both an observation slice and a ``direction candidate'' for pruning.

\textbf{Candidate pruning.} A candidate retriever outputs a subset of direction indices $\mathcal{I}_t \subseteq \{1,\dots,K\}$ (e.g., $|\mathcal{I}_t|=5$), which induces a pruned observation:
\begin{equation}
    \tilde{o}_t = (s_t^{scene}, \{ d_{t,k} \}_{k \in \mathcal{I}_t}),
\end{equation}
If we also expose an explicit navigable list, it is correspondingly pruned to $\tilde{\mathcal{V}}_t = \bigcup_{k \in \mathcal{I}_t} \mathcal{V}_{t,k}$.

\textbf{Action space.} The agent selects an action $a_t$ from the action space:
\begin{equation}
\mathcal{A}_t = \mathcal{V}_t \cup \{\texttt{finished}\},
\end{equation}
where selecting a viewpoint ID moves the agent to that viewpoint, and \texttt{finished} terminates the episode.

\textbf{History.} We denote the navigation history up to step $t$ as:
\begin{equation}
h_t = \{(a_0, o_0), (a_1, o_1), \ldots, (a_{t-1}, o_{t-1})\},
\end{equation}
which is stored and fed to the LLM as a textual scratchpad summarizing prior actions and observations.

\textbf{State.} We define the agent state as the tuple:
\begin{equation}
s_t = (I, o_t, h_t),
\end{equation}
i.e., the instruction, the current textual observation, and the navigation history.

\textbf{Policy and trajectory.} A navigation policy $\pi$ maps the state to an action:
\begin{equation}
a_t \sim \pi(\cdot \mid s_t),
\end{equation}
The executed actions produce a trajectory $\tau = (v_0, a_0, v_1, \ldots, a_{T-1}, v_T)$.

\textbf{Success criterion and metrics.} Following standard R2R evaluation, an episode is successful if the agent terminates with \texttt{finished} at a final viewpoint $v_T$ whose shortest-path distance to the goal viewpoint is less than 3 meters. We evaluate using Success Rate (SR) and Success weighted by Path Length (SPL), and additionally report efficiency metrics relevant to LLM-based navigation.

\section{Methodology}

\begin{figure*}[t]
    \centering
    \includegraphics[width=0.8\textwidth]{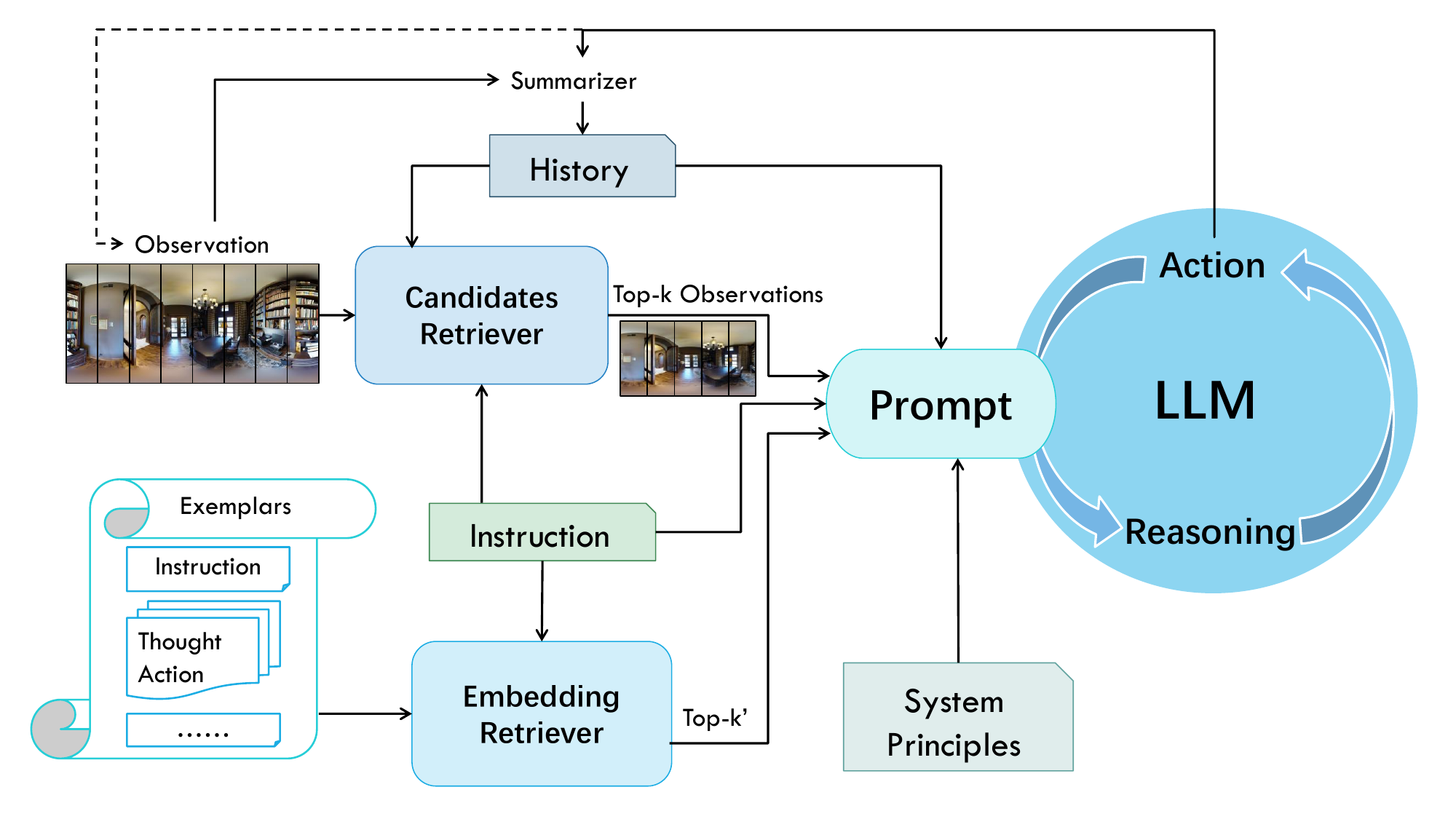}
    \caption{The agent follows a language-centric pipeline. At each step, a candidate retriever prunes directional observations before LLM reasoning, while an embedding-based instruction retriever retrieves exemplar trajectories at the episode level for in-context guidance. A history summarizer maintains a compact navigation context. The prompt aggregates the system principles, instruction, retrieved exemplars, pruned observations, and history, enabling the LLM to generate reasoning traces and navigation actions in a closed-loop manner.}
    \label{fig:arch}
    \vspace{-10pt}
\end{figure*}

We present a retrieval-augmented LLM navigator that extends a NavGPT-style baseline \cite{zhou2023navgpt} with two retrieval modules: (i) an embedding-based demonstration retriever for in-context learning (ICL) and (ii) an imitation-learned candidate retriever for top-$k$ directional pruning.

\subsection{Model Architecture}
\label{sec:baseline}
We adopt a language-centric navigation architecture that performs sequential decision making by combining instruction, textualized perception, and navigation history \cite{zhou2023navgpt}. At each step $t$, the agent is located at viewpoint $v_t$ and represents the panoramic sensory input as a textual observation $o_t=(s_t^{scene},\{d_{t,k}\}_{k=1}^{8})$, where each $d_{t,k}$ summarizes the scene and nearby objects within a directional sector and lists the navigable viewpoint IDs in that sector. A prompt manager $\mathcal{M}$ aggregates the instruction $I$, current observation $o_t$, and history buffer $h_t$ into a single prompt, which is then consumed by an LLM (Qwen3\cite{yang2025qwen3} in our baseline) to produce an explicit reasoning trace $r_t$ and the next action $a_t\in\mathcal{A}_t$. Executing viewpoint-ID actions transitions the agent to $v_{t+1}$, after which the new observation and the reasoning--action trace are appended to $h_{t+1}$ to support progress tracking and long-horizon planning.

On top of this baseline, we introduce two lightweight yet complementary retrieval modules to support LLM reasoning. As illustrated in Fig.~\ref{fig:arch}, an \emph{instruction-level} retriever operates once at the beginning of each navigation episode, retrieving exemplar trajectories that are semantically similar to the current instruction $I$. These exemplars are used to augment the prompt and provide high-level reasoning references. During navigation, a \emph{candidate retriever} is invoked at every time step to score and prune the directional candidate set $\mathcal{D}_t=\{d_{t,k}\}_{k=1}^{8}$ given the current state $s_t=(I,o_t,h_t)$, producing a subset of indices $\mathcal{I}_t$ and a pruned observation $\tilde{o}_t=(s_t^{scene},\{d_{t,k}\}_{k\in\mathcal{I}_t})$. The filtered candidates are then passed to the LLM, which produces the final navigation decision in the original action format. This design preserves the core decision pipeline while explicitly injecting task-relevant priors and reducing unnecessary reasoning over irrelevant candidates.

\subsection{Embedding-based Demonstration Retriever}
\label{sec:demo-retriever}
LLM navigators are sensitive to the choice of in-context demonstrations. Rather than using a single static demonstration, we dynamically retrieve relevant successful trajectories conditioned on the episode instruction. To enable instruction-level retrieval, we maintain an exemplar memory $\mathcal{E}$ from successful navigation trajectories stored in a static file, where each exemplar $e\in\mathcal{E}$ contains (i) a task instruction string and (ii) a step-wise navigation trace formatted in the same \texttt{Thought/Action/Observation} style as the navigator. Each exemplar is also associated with a precomputed embedding vector, allowing efficient similarity search at runtime. This memory serves as a repository of prior navigation experiences, allowing the agent to reuse previously successful reasoning patterns when encountering semantically similar instructions.

\textbf{Embedding and similarity.} Given a new navigation episode, the input instruction is encoded into a fixed-dimensional embedding using a pretrained text encoder. All exemplar instructions in the memory are encoded in the same embedding space. 
Let $f_{\phi}(\cdot)$ be a pretrained sentence embedding model. Given the episode instruction $I$, we compute a query embedding $q=f_{\phi}(I)$ and rank exemplars by cosine similarity to the stored exemplar embedding $z_e$:
\begin{equation}
    \mathrm{sim}(I,e)=\frac{\langle q, z_e\rangle}{\|q\|\,\|z_e\|},
\end{equation}
We select the top-$k$ demonstrations $\mathcal{E}_k(I)$ and format them into an \texttt{examples} block.

\textbf{Prompt construction.} The \texttt{examples} block is prepended to the navigation prompt once per episode (before step $t=0$). Each exemplar includes the instruction and its corresponding trajectory summary, presented as a reference navigation experience. This provides instruction-conditioned guidance on (i) how to interpret directional observations, (ii) how to select viewpoint IDs, and (iii) when to output \texttt{finished}, improving stability without modifying the LLM weights.
Importantly, exemplars provide \emph{soft} in-context guidance rather than templates to be strictly imitated. The LLM may deviate from exemplar behaviors when the current observation differs, preserving flexibility in unseen environments.

\subsection{Imitation-learned Candidate Retriever}
\label{sec:cand-retriever}
In language-driven navigation, the LLM is often required to reason over verbose candidate descriptions at every step, which increases inference cost and can amplify errors from irrelevant or distracting options. We introduce a lightweight candidate retriever that prunes the directional candidate set $\mathcal{D}_t=\{d_{t,k}\}_{k=1}^{8}$ before LLM reasoning, reducing prompt length while preserving valid navigation options.

\subsubsection{State-Action Representation}

At each step $t$, the candidate retriever operates on the same information available to the navigator. We build a context string:
\begin{equation}
    c_t = \mathrm{Concat}(I, h_t),
\end{equation}
and use the 8 directional descriptions $\{d_{t,k}\}_{k=1}^{8}$ (each containing a direction-specific scene description, nearby objects, and the viewpoint IDs in that sector) as candidates.
We encode the context and each directional candidate with a sentence encoder $g_{\psi}(\cdot)$:
\begin{equation}
    u_t=g_{\psi}(c_t), \qquad z_{t,k}=g_{\psi}(d_{t,k}),
\end{equation}
The relative spatial cues are included implicitly in the textual descriptions $d_{t,k}$. We score each direction independently with a lightweight MLP head:
\begin{equation}
    \alpha_{t,k}=\mathrm{MLP}_{\theta}([u_t;z_{t,k}]),
\end{equation}
We then select the top-$k$ direction indices $\mathcal{I}_t=\mathrm{TopK}(\{\alpha_{t,k}\}_{k=1}^{8})$ and pass the pruned observation $\tilde{o}_t=(s_t^{\mathrm{scene}},\{d_{t,k}\}_{k\in\mathcal{I}_t})$ to the LLM navigator.

\subsubsection{Training via Imitation Learning}
We train the retriever with imitation learning on R2R trajectories \cite{AndersonWTB0S0G18}. For each training trajectory and each step $t$, we construct the 8 directional descriptions with the same formatting used at inference time, and label the ground-truth direction $y_t \in \{1,\dots,8\}$ that contains the next viewpoint on the reference path.
We optimize cross-entropy over the 8 directions:
\begin{equation}
    \mathcal{L}_{\text{ret}}=-\sum_t \log \frac{\exp(\alpha_{t,y_t})}{\sum_{k=1}^{8}\exp(\alpha_{t,k})},
\end{equation}

In practice, $g_{\psi}$ is instantiated with a pretrained sentence transformer encoder, while the MLP head is trained to predict the correct direction.
Training is performed offline and does not involve updating the LLM parameters.
This separation allows the retriever to learn task-specific priors while keeping the language model frozen.

\subsubsection{Inference-time Integration}
Each selected direction $k\in\mathcal{I}_t$ contains a set of navigable viewpoint IDs $\mathcal{V}_{t,k}$. During inference, the candidate retriever scores all navigable actions at the current time step and selects the top-$k$ candidates.
Only these directions are included in the prompt passed to the LLM. Consequently, the LLM effectively chooses the next viewpoint ID from $\tilde{\mathcal{V}}_t=\bigcup_{k\in\mathcal{I}_t}\mathcal{V}_{t,k}$, which reduces candidate noise and improves inference efficiency.
This integration reduces prompt length and inference cost without modifying the LLM’s reasoning mechanism.

\begin{table*}[ht]
\begin{tabular}{l|ccccc|ccccc}
\toprule
\multirow{2}{*}{ \ \ \ Methods}  &\multicolumn{5}{c}{Val Seen}&\multicolumn{5}{c}{Val Unseen}\\ & 
\multicolumn{1}{c}{TL} &
\multicolumn{1}{c}{NE$\downarrow$} & 
\multicolumn{1}{c}{SR$\uparrow$} & 
\multicolumn{1}{c}{OSR$\uparrow$} & 
\multicolumn{1}{c}{SPL$\uparrow$}
 & TL & NE$\downarrow$ & SR$\uparrow$ & OSR$\uparrow$ & SPL$\uparrow$ \\
 \midrule
 
 \ \ \ Seq2Seq \cite{AndersonWTB0S0G18} 
 & 11.33 & 6.01 & 39 & 53 & - 
 & 8.39 & 7.81 & 21 & 28 & - \\
 
 \ \ \ SpeakerFollower \cite{fried2018speakerfollower} 
 & - & 3.36 & 66 & 74 & - 
 & - & 6.62 & 36 & 45 & - \\
 
 \ \ \ HAMT \cite{Chen2021HAMT} 
 & 11.15 & 2.52 & 76 & - & 72 
 & 11.46 & 2.29 & 66 & - & 61 \\
 
 \ \ \ DUET \cite{Chen2022DUET} 
 & 12.32 & 2.28 & 79 & 86 & 73 
 & 13.94 & 3.31 & 72 & 81 & 60 \\
 
 \ \ \ BEVBert \cite{an2023bevbert} 
 & 13.56 & 2.17 & 81 & \textbf{88} & 74 
 & 14.55 & 2.81 & 75 & 84 & 64 \\
 
 \ \ \ ScaleVLN \cite{wang2023scalingdata} 
 & 13.24 & \textbf{2.12} & \textbf{81} & 87 & \textbf{75} 
 & 14.09 & \textbf{2.09} & \textbf{81} & \textbf{88} & \textbf{70} \\
 
 \ \ \ NavGPT \cite{zhou2023navgpt} 
 & - & - & - & - & - 
 & 11.45 &6.46 &34 & 42 & 29 \\

\midrule

 \ \ \ NavGPT (Qwen-3) 
 & 15.29 & 8.81 & 15.77 & 29.87 & 10.30 
 & 15.66 & 8.27 & 18.22 & 33.25 & 11.40 \\
 
\rowcolor{blue!7}  \ \ \ \textbf{Our Method} 
& 16.20 & 8.96 & 19.88 & 39.86 & 13.29 
& 16.35 & 8.33 & 23.41 & 44.70 & 14.76 \\ 

 \bottomrule
\end{tabular}
\centering
\caption{Comparison with the state of the art on R2R. \textbf{Bold}  highlight the best and runner-up performance in each column, while \colorbox{blue!10}{blue} underscore our method row. $\uparrow$ indicates better performance with higher values.}
% , whereas $\downarrow$ indicates better performance with lower values.}
\label{tab:r2r-sota}
\end{table*}

\begin{table*}[t]
\begin{tabular}{l l | cc ccc | cc ccc | c}
\toprule
\multirow{2}{*}{Setting} & \multirow{2}{*}{Method} 
& \multicolumn{5}{c|}{Val Seen} 
& \multicolumn{5}{c}{Val Unseen} 
& \multicolumn{1}{c}{Tokens / Time}\\
% \cmidrule(lr){3-7} \cmidrule(lr){8-12}
 & 
 & TL 
 & NE$\downarrow$ 
 & SR$\uparrow$ 
 & OSR$\uparrow$ 
 & SPL$\uparrow$ 
 & TL 
 & NE$\downarrow$ 
 & SR$\uparrow$ 
 & OSR$\uparrow$ 
 & SPL$\uparrow$
 & (per episode) \\
\midrule

% \multirow{6}{*}{cross-modal backbone}
\multirow{6}{*}{\makecell[l]{Ablation\\Mode}}

& Baseline 
& 15.29 & 8.81 & 15.77 & 29.87 & 10.30 
& 15.66 & 8.27 & 18.22 & 33.25 & 11.40 & 24.8k / 17.9s \\

& Baseline + Embedding Retriever 
& 16.68 & 9.21 & 17.04 & 33.99 & 10.84 
& 16.83 & 8.39 & 20.14 & 38.95 & 12.54 & 35.0k / 9.76s \\

& Baseline + Candidate Retriever
& 15.18 & 8.75 & 21.65 & 36.14 & 14.03 
& 15.47 & 8.28 & 20.31 & 37.89 & 12.97 & 22.1k / 16.1s \\

& Full Mode + Thinking Mode$^*$ 
& 14.93 & \textbf{7.91} & \textbf{24.00} & 39.00 & \textbf{15.91} 
& 14.11 & \textbf{8.02} & \textbf{26.40} & 44.40 & \textbf{18.00} & 57.6k / 536.5s \\

& Full Mode (Ours)
& 16.20 & 8.96 & 19.88 & \textbf{39.86} & 13.29 
& 16.35 & 8.33 & 23.41 & \textbf{44.70} & 14.76 & 32.1k / 10.1s \\

\midrule

\multirow{4}{*}{\makecell[l]{LLM\\Backbone}}
& LLaMA 3.1 \cite{grattafiori2024llama3}
& 13.80 & 8.89 & 10.38 & 21.65 & 6.32 
& 13.97 & \textbf{8.24} & 13.18 & 24.96 & 8.50 & 35.4k / 36.1s \\

& Qwen 3 \cite{yang2025qwen3} 
& 16.20 & 8.96 & 19.88 & 39.86 & \textbf{13.29} 
& 16.35 & 8.33 & \textbf{23.41} & 44.70 & \textbf{14.76}
& 32.1k / 10.1s \\

& GPT 4o mini \cite{openai2024gpt4}
& 19.80 & \textbf{8.67} & \textbf{22.43} & \textbf{48.78} & 11.64 
& 19.49 & 8.43 & 22.78 & \textbf{56.24} & 11.60 & 30.1k / 14.4s\\

\bottomrule
\end{tabular}
\centering
\caption{Ablation Studies. Thinking Mode$^*$ is enabled by using a reasoning-capable model with its internal “thinking” mode turned on. Unless otherwise specified, all baseline results are based on NavGPT with Qwen3-8B \cite{yang2025qwen3} as the backbone model. $\uparrow$ indicates better performance with higher values.}
% , whereas $\downarrow$ indicates better performance with lower values.}
% * indicates results obtained by randomly sampling 10\% of the R2R dataset.}
\label{tab:ablation}
\end{table*}

\section{Experiment}

\subsection{Research Questions}
Our experiments are designed to answer the following research questions (RQs):
 
% \begin{itemize}
%     \item \textbf{RQ1: Efficiency.}
%     How does retrieval augmentation affect inference cost and navigation performance in LLM-based VLN?
%     We analyze prompt length, inference time, and navigation metrics to assess the trade-off between computational overhead and performance gains.
%     \item \textbf{RQ2: CoT Reasoning Mode.}  
%     Does explicit chain-of-thought reasoning improve navigation performance, and is it practical under realistic resource constraints? 
%     \item \textbf{RQ3: Qualitative Behavior.}  
%     How does retrieval augmentation influence navigation decisions in representative success cases?
% \end{itemize}
\begin{itemize}
    % \item \textbf{RQ1: Overall Performance.}  
    \item \textbf{RQ1:} Does retrieval augmentation improve navigation performance in LLM-based VLN?
    % We evaluate the proposed framework on the R2R benchmark and compare it with an LLM-only baseline in terms of Success Rate (SR), Oracle Success Rate (OSR), and SPL on both seen and unseen environments.

    % \item \textbf{RQ2: Component Effectiveness.}  
    \item \textbf{RQ2:} What are the individual and combined contributions of instruction-level exemplar retrieval and step-level candidate pruning?
    % We conduct ablation studies to analyze how each retrieval module affects navigation success and trajectory efficiency.

    % \item \textbf{RQ3: Efficiency and Inference Cost.}  
    \item \textbf{RQ3:} How does retrieval augmentation affect inference efficiency in LLM-based navigation?
    % We analyze prompt length, inference time, and computational overhead to assess the trade-off between efficiency and performance gains.

    \item \textbf{RQ4:}
    %Qualitative Navigation Behavior.}  
    How does retrieval augmentation influence navigation decisions and error patterns in representative navigation episodes?
    % We provide qualitative case studies to illustrate how exemplar retrieval and candidate pruning affect instruction grounding and step-wise decision making.
\end{itemize}

\subsection{Experiment Setup}

\subsubsection{Dataset}
We evaluate our method on the Room-to-Room (R2R) benchmark~\cite{AndersonWTB0S0G18}, which is built upon Matterport3D indoor environments.
R2R consists of 7,189 navigation trajectories across 90 real-world scenes, each paired with three natural-language instructions.
Agents navigate on a discrete viewpoint graph and are required to follow the instructions to reach a target location. We report results on the standard \texttt{Val Seen} and \texttt{Val Unseen} splits, containing 1,021 and 2,349 instruction–trajectory pairs, respectively.
Following the standard protocol, an episode is considered successful if the agent terminates within 3 meters of the goal viewpoint (Please refer to Section~\ref{sec:problem}).

\subsubsection{Baselines}
\begin{figure}[t]
    \centering
    \includegraphics[width=1\linewidth]{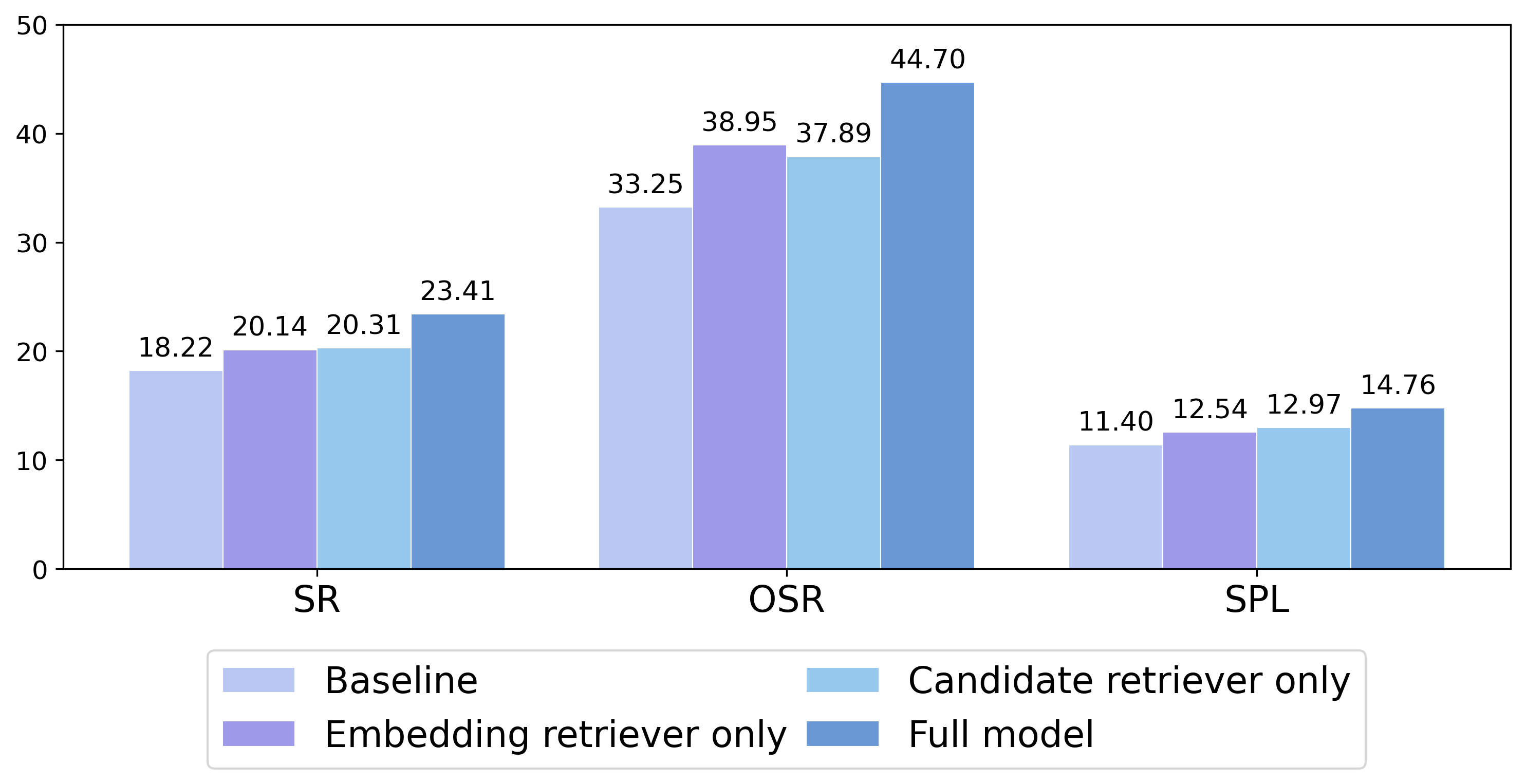}
    \caption{SR, OSR, and SPL of the baseline agent with different ablation modules on the R2R val-unseen split.}
    \label{fig:abl}
    \vspace{-10pt}
\end{figure}

Our primary baseline is a \textit{Qwen3-based} implementation of the NavGPT navigation pipeline \cite{zhou2023navgpt}. This baseline introduces no algorithmic modifications. We only replace the original LLM used in NavGPT with Qwen3, while keeping all other components unchanged, including the prompt format, the 8-direction textual observation representation, the navigation rules, and the interaction loop with the simulator. 
We compare this Qwen3-based NavGPT baseline against our full retrieval-augmented system. Fig.~\ref{fig:abl} further provides a bar-chart comparison between the baseline and our method, highlighting consistent gains in SR/OSR/SPL across splits.

\subsubsection{Evaluation Metrics}
We adopt standard R2R evaluation metrics \cite{zhou2023navgpt}.
\textbf{TL} (Trajectory Length) measures the number of executed actions (viewpoint transitions) before termination.
\textbf{NE} (Navigation Error) computes the shortest-path distance (in meters) between the final viewpoint and the goal viewpoint, where lower values indicate better goal localization.
\textbf{SR} (Success Rate) ireports the fraction of episodes with $\mathrm{NE} < 3$m at termination.
\textbf{OSR} (Oracle Success Rate) is the fraction of episodes where any visited viewpoint along the trajectory comes within 3 meters of the goal, which measures whether the agent can reach the vicinity of the target even if it fails to stop optimally.
\textbf{SPL} (Success weighted by Path Length) measures efficiency by weighting success with the ratio between the shortest-path length and the executed path length, penalizing unnecessary detours. Higher SPL reflects both higher success and shorter trajectories.

\subsubsection{Implementation Details}
% For all baselines, we carefully tune their hyperparameters to achieve the best performance on the validation set.  
For our method, we adopt Qwen-3 (8B) \cite{yang2025qwen3} as the LLM backbone and Qwen embedding (0.6B) \cite{zhang2025qwen3emb} for instruction-level retrieval.
The candidate retriever is trained using a sentence-transformer encoder all-MiniLM-L6-v2 \cite{reimers2019sentencebertsentenceembeddingsusing} followed by an MLP scoring head.
All experiments are conducted on a single NVIDIA A100 GPU.
The candidate retriever is trained offline for $10$ epochs using the R2R training split, with a batch size of $32$, AdamW optimizer \cite{LoshchilovH19}, and a learning rate of $2\times10^{-5}$.
We supervise the retriever using shortest-path labels over $8$ navigable directions.
% and obtain a Recall@5 of $0.9918$ on the validation unseen split (Note that the retriever predicts directional sectors rather than exact viewpoints, which makes the classification task easier but sufficient for effective pruning.).
During inference, we retrieve the top-3 instruction exemplars and top-5 candidate actions by default.
The exemplar memory $\mathcal{E}$ for the embedding retriever contains $20$ demonstration samples. 
% from \texttt{Val Unseen}.

\begin{figure*}[ht]
    \centering
    \includegraphics[width=0.85\textwidth]{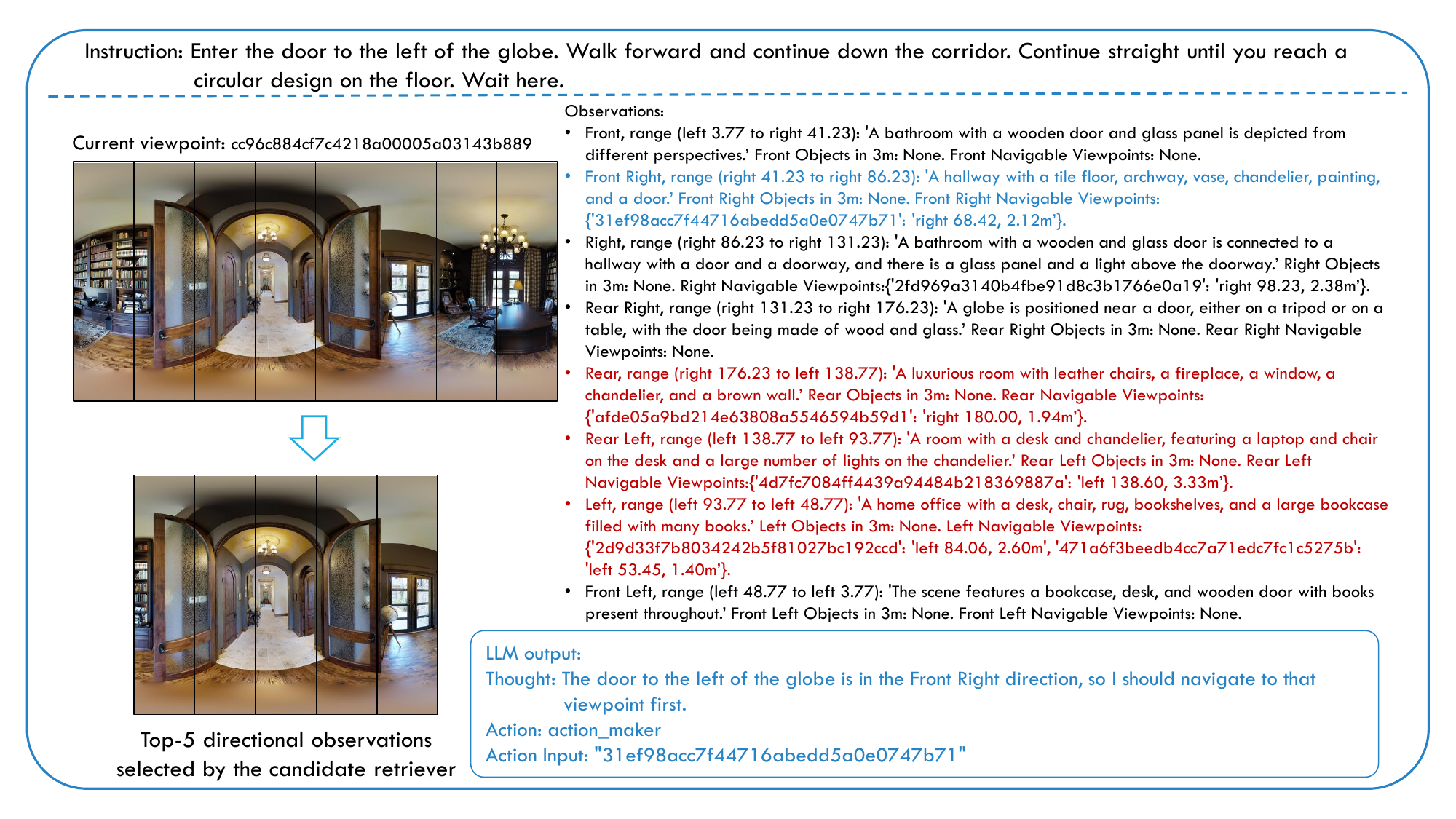}
    \caption{Here is a case study of a successful navigation step. At this step, the candidate retriever selects the top 5 directional observations for LLM reasoning. The direction highlighted in blue corresponds to the correct navigable direction chosen by the LLM, while the red directions indicate irrelevant candidates pruned by the candidate retriever. This pruning reduces decision noise and enables the LLM to focus on instruction-consistent actions.}
    \label{fig:case}
    \vspace{-10pt}
\end{figure*}

\subsection{Main Result (RQ1)}

Table~\ref{tab:r2r-sota} summarizes the main results on the R2R benchmark.
We compare our retrieval-augmented method with both classical supervised VLN agents and recent LLM-based navigation approaches. As expected, fully supervised and pretraining-based methods, such as HAMT\cite{Chen2021HAMT}, DUET\cite{Chen2022DUET}, and ScaleVLN\cite{wang2023scalingdata}, achieve substantially higher absolute performance due to large-scale task-specific training and stronger cross-modal architectures. In contrast, our approach keeps the navigator LLM frozen and introduces only lightweight retrieval modules, targeting practical improvements within an LLM-centric pipeline.
Compared to the Qwen3-based baseline, our method improves Success Rate (SR) from 15.77\% to 19.88\% on \texttt{Val Seen}, and from 18.22\% to 23.41\% on \texttt{Val Unseen}.
These gains are accompanied by consistent improvements in Oracle Success Rate (OSR), which increases from 29.87\% to 39.86\% on \texttt{Val Seen} and from 33.25\% to 44.70\% on \texttt{Val Unseen}.
Similarly, SPL improves from 10.30 to 13.29 on \texttt{Val Seen} and from 11.40 to 14.76 on \texttt{Val Unseen},  indicating that retrieval augmentation not only increases the likelihood of reaching the goal region, but also encourages more efficient trajectories.
Navigation Error (NE) remains comparable, suggesting that improvements primarily come from better step-wise decisions rather than simply shorter paths. Notably, gains on \texttt{Val Unseen} is more pronounced, suggesting that retrieval provides stronger benefits under distribution shift. In terms of efficiency, in our profiling, our method processes approximately 32.1k input tokens per episode and requires an average inference time of 10.1 seconds per episode. While retrieval introduces additional computation, this overhead is partially offset by improved success and path efficiency, as well as reduced reasoning over irrelevant candidates.

Overall, these results demonstrate that retrieval augmentation is an effective and scalable strategy for improving LLM-based vision-language navigation, narrowing the performance gap with supervised methods while preserving the flexibility and transparency of language-driven decision making.

\subsection{Ablation Study (RQ2 and RQ3)}

Table~\ref{tab:ablation} presents a detailed ablation study that analyzes the contribution of each component in our retrieval-augmented framework, as well as the impact of different LLM backbones and reasoning modes.

\textbf{Effect of the instruction-level embedding retriever.}
Adding only the instruction-level embedding retriever leads to consistent improvements over the baseline on both splits.
On \texttt{Val Unseen}, the Success Rate increases from 18.22\% to 20.14\%, and Oracle Success Rate improves notably from 33.25\% to 38.95\%.
This suggests that retrieving instruction-level exemplars provides useful global guidance, helping the LLM better interpret task intent and maintain progress toward the goal.
However, the improvement in SPL is comparatively smaller, (11.40 $\rightarrow$ 12.54), indicating that exemplar retrieval alone does not sufficiently address inefficiencies arising from step-level action ambiguity.

\textbf{Effect of the candidate retriever.}
In contrast, adding only the candidate retriever yields more substantial gains in trajectory efficiency.
On \texttt{Val Seen}, SPL improves from 10.30 to 14.03, while SR increases from 15.77\% to 21.65\%.
Similar trends are observed on \texttt{Val Unseen}.
These results support the hypothesis that pruning irrelevant navigable candidates effectively reduces local decision noise, allowing the LLM to focus on a smaller, more relevant action set.
Compared with exemplar retrieval, candidate pruning has a stronger impact on SPL, highlighting its role in improving step-wise efficiency rather than global planning.

\textbf{Complementarity of both retrieval modules.}
Combining both retrievers yields consistent improvements across all metrics compared to either module alone.
Our full model achieves higher OSR on both splits (39.86\% on \texttt{Val Seen}, 44.70\% on \texttt{Val Unseen}), indicating improved ability to reach goal-adjacent regions.
At the same time, SR and SPL also improve over the baseline, confirming that global instruction-level priors and local candidate pruning are complementary.
% Exemplar retrieval provides high-level reasoning references at the episode level, while the candidate retriever reduces ambiguity during per-step decision making.
From an efficiency perspective, although exemplar augmentation increases the prompt length (from 24.8k to 32.1k tokens per episode), the full model reduces inference time from 17.9s to 10.1s per episode.
This indicates that exemplar retrieval provides high-level reasoning references at the episode level, while the candidate retriever reduces ambiguity during per-step decision-making.

% \textbf{Chain of thought reasoning mode analysis.}
% We further evaluate the effect of enabling explicit chain-of-thought (CoT) reasoning on top of the full retrieval framework.
% As shown in Table~\ref{tab:ablation}, the \emph{full mode + thinking mode} configuration achieves the strongest performance among all ablation settings.
% On \texttt{Val Unseen}, SR increases from 23.41\% to 26.40\%, and SPL improves substantially from 14.76 to 18.00.
% Navigation Error is also reduced (8.33 $\rightarrow$ 8.02), suggesting more coherent long-horizon planning.
% These gains indicate that explicit reasoning traces help the LLM maintain consistent plans and resolve ambiguities introduced by long navigation histories.
% However, this improvement comes at a high computational cost. In our experiments, enabling reasoning mode increased generation length and inference latency dramatically, requiring approximately 12 hours to evaluate only 150 episodes. Due to these resource constraints, we do not adopt the reasoning mode in our default setting and treat it as an optional accuracy-boosting configuration when computational budget permits.
% Therefore, the results reported in Table~\ref{tab:ablation} with reasoning mode enabled are obtained by randomly sampling 10\% of the episodes from both the \texttt{Val Seen} and \texttt{Val Unseen} splits.
% This setting substantially reduces evaluation time and computational overhead, while random sampling provides a reasonable approximation of the performance trends on the full dataset.

\textbf{Chain-of-thought reasoning mode analysis.}
We evaluate the effect of enabling explicit CoT reasoning on top of the full retrieval framework. As shown in Table~\ref{tab:ablation}, the \emph{full mode + thinking mode} configuration achieves the strongest performance among all ablation settings.
On \texttt{Val Unseen}, SR increases from 23.41\% to 26.40\%, SPL improves from 14.76 to 18.00, and Navigation Error is reduced (8.33 $\rightarrow$ 8.02), indicating more coherent long-horizon planning.
These results suggest that explicit reasoning traces help the LLM maintain consistent plans and resolve ambiguities from long navigation histories.
However, CoT reasoning incurs substantial computational overhead: enabling reasoning mode significantly increases generation length and inference latency, requiring approximately 12 hours to evaluate 150 episodes.
As a result, we do not adopt the reasoning mode in the default setting and treat it as an optional accuracy-boosting configuration.
Due to resource constraints, the reported results with reasoning mode are obtained by randomly sampling 10\% of episodes from both the \texttt{Val Seen} and \texttt{Val Unseen} splits, which provides a reasonable approximation of overall performance trends.

\textbf{LLM backbone comparison.}
Finally, we evaluate different LLM backbones under the same retrieval framework.
GPT-4o mini \cite{openai2024gpt4} achieves strong oracle success, particularly on \texttt{Val Unseen} (OSR 56.24\%), but exhibits lower SPL due to longer and less efficient trajectories.
LLaMA~3.1 \cite{grattafiori2024llama3} underperforms across most metrics, especially in SR and SPL, suggesting weaker instruction grounding in this setting.
Qwen~3 \cite{yang2025qwen3} achieves a favorable balance between success rate, trajectory efficiency, and computational cost, making it a practical choice for our framework.
These results demonstrate that our retrieval-augmented design generalizes across LLM backbones and is not tied to a specific model.

\subsection{Qualitative Analysis (RQ4)}

We qualitatively analyze a representative navigation step in Fig.~\ref{fig:case} to illustrate how retrieval-augmented reasoning improves decision making. In this example, the candidate retriever prunes visually plausible but instruction-irrelevant directional observations (marked in red), reducing prompt clutter and constraining the LLM’s reasoning to a small set of instruction-consistent candidates. Conditioned on this pruned observation set, the LLM correctly grounds the phrase “to the left of the globe” to the appropriate front-right navigable viewpoint (highlighted in blue) and selects the correct action. This case highlights the complementary roles of the two retrieval components: instruction-level exemplars provide high-level guidance on navigation intent, while candidate pruning enforces local action relevance, together enabling more focused reasoning and improved robustness in complex indoor environments.
\section{Conclusion}

% We presented a retrieval-augmented LLM navigation framework that improves a Qwen3-based VLN navigator without modifying the underlying language model. Our method introduces two complementary modules: an instruction-level embedding retriever that injects relevant exemplar trajectories as in-context guidance, and an imitation-learned candidate retriever that prunes 8-direction observations to a small set of likely directions before LLM reasoning. Experiments on R2R show consistent gains over the LLM-only baseline, improving SR/OSR/SPL on both seen and unseen splits while reducing unnecessary reasoning over irrelevant candidates.

% Despite these improvements, our approach remains limited by the fidelity of text-only observations, potential domain shift in exemplar and candidate retrieval, and the computational cost of long-context or thinking-style inference. Future work includes integrating stronger multimodal perception, learning retrieval objectives that better align with downstream navigation success, and combining retrieval with verification or lightweight fine-tuning to further close the gap to fully trained VLN agents.

% Despite these gains, the approach is limited by text-only observations, potential domain shift in retrieval, and the cost of long-context or reasoning-style inference. 

In this paper, we propose a retrieval-augmented LLM navigation framework that improves the VLN agents without modifying the underlying language model. By combining an instruction-level exemplar retriever for in-context guidance and an imitation-learned candidate retriever for directional pruning, our method achieves consistent improvements in SR, OSR, and SPL on both seen and unseen splits of the R2R benchmark while reducing reasoning over irrelevant candidates. 
Future work will explore stronger multimodal perception, retrieval objectives better aligned with navigation success, and tighter integration with verification or lightweight fine-tuning.

\bibliographystyle{IEEEtran}
\bibliography{ref}

\end{document}